\documentclass[10pt,twocolumn,letterpaper]{article}

\usepackage{cvpr}       % To produce the REVIEW version

\usepackage[table]{xcolor}
\usepackage{capt-of}
\usepackage{xspace}
\usepackage{xcolor}
\usepackage{enumitem}
\usepackage{amsmath,amsthm,amssymb,amsfonts,dsfont,pifont,bm,bbm,mathrsfs,mathtools,nicefrac,extarrows,relsize}
\usepackage{algorithm,algpseudocode,listings}
\usepackage{booktabs,multirow,adjustbox,diagbox,threeparttable,tabularray,setspace}

\definecolor{cvprblue}{rgb}{0.21,0.49,0.74}
\usepackage[pagebackref,breaklinks,colorlinks,citecolor=cvprblue,bookmarks=false]{hyperref}
\usepackage{wrapfig}
\usepackage[capitalize]{cleveref}  % Should be loaded after 'hyperref', and works perfectly with 'subfigure'.

\crefname{section}{Sec.}{Secs.}
\Crefname{section}{Section}{Sections}
\crefname{appendix}{App.}{Apps.}
\Crefname{appendix}{Appendix}{Appendices}
\crefname{table}{Tab.}{Tabs.}
\Crefname{table}{Table}{Tables}
\crefname{figure}{Fig.}{Figs.}
\Crefname{figure}{Figure}{Figures}
\crefname{equation}{Eq.}{Eqs.}
\Crefname{equation}{Equation}{Equations}
\crefname{theorem}{Thm.}{Thms.}
\Crefname{theorem}{Theorem}{Theorems}
\crefname{lemma}{Lem.}{Lems.}
\Crefname{lemma}{Lemma}{Lemmas}
\crefname{remark}{Rem.}{Rems.}
\Crefname{remark}{Remark}{Remarks}
\crefname{corollary}{Cor.}{Cors.}
\Crefname{corollary}{Corollary}{Corollaries}
\crefname{algorithm}{Alg.}{Algs.}
\Crefname{algorithm}{Algorithm}{Algorithms}
\definecolor{cellred}{RGB}{213, 123, 101}
\definecolor{cellgreen}{RGB}{0, 205, 0}
\definecolor{cellblue}{RGB}{54, 125, 189}
\definecolor{codegreen}{rgb}{0,0.6,0}
\definecolor{codegray}{rgb}{0.5,0.5,0.5}
\definecolor{codepurple}{rgb}{0.58,0,0.82}
\definecolor{backcolour}{rgb}{1.0,1.0,1.0}
\lstdefinestyle{mystyle}{
    backgroundcolor=\color{backcolour},
    commentstyle=\color{codegreen},
    keywordstyle=\color{magenta},
    numberstyle=\tiny\color{codegray},
    stringstyle=\color{codepurple},
    basicstyle=\ttfamily\scriptsize,
    breakatwhitespace=false,
    breaklines=true,
    captionpos=b,
    keepspaces=true,
    numbers=left,
    numbersep=5pt,
    showspaces=false,
    showstringspaces=false,
    showtabs=false,
    tabsize=2
}
\newcommand{\tocite}[1]{{\color{red} [TO CITE]}}

\newcommand{\name}{\texttt{\textbf{MovieGrid}}\xspace}

\usepackage[most]{tcolorbox}
\usepackage{enumitem}
\usepackage{xcolor}

\definecolor{promptbg}{RGB}{249,249,249}
\definecolor{promptborder}{RGB}{205,205,205}

\newtcolorbox{promptbox}{
    enhanced,
    colback=promptbg,
    colframe=promptborder,
    boxrule=0.6pt,
    arc=3pt,
    outer arc=3pt,
    left=12pt,
    right=12pt,
    top=12pt,
    bottom=12pt,
    boxsep=0pt,
    width=\linewidth,
}

\def\confName{CVPR}
\def\confYear{2026}

\title{Multi-Grid Post-Training for Long-Form Multi-Shot Video Generation}

\author{Jiawei Mao$^{1}$ \, \, 
  Haoqin Tu$^{1}$ \, \,
  Hardy Chen$^{1}$ \, \,
  Yuhan Wang$^{1}$ \, \, 
  Keyang Xu $^{4}$ \, \, 
  Jieru Mei $^{4}$ \vspace{.3em}\\
  Hongliang Fei $^{4}$ \, \,
  Ruogu Fang$^{2,3}$ \, \,
  Wei Shao$^{2}$ \, \,
  Cihang Xie$^{1}$ \, \,
  Yuyin Zhou$^{1}$ \vspace{.5em}\\ 
 $^{1}$ UC Santa Cruz \, \, $^{2}$ University of Florida \, \, $^{3}$ Vanderbilt University \, \, $^{4}$ Google
\vspace{.5em}
  \\
  \small
  \hspace{3em} \includegraphics[height=1.1em]{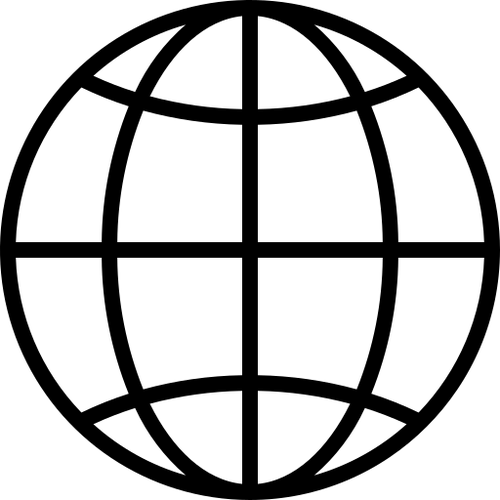} \textbf{Project Page}: \url{https://jwmao1.github.io/moviegrid_web} \\
  \vspace{-2.5em}}

\makeatletter\let\ps@plain\ps@empty\makeatother
\begin{document}
\twocolumn[{
    \renewcommand\twocolumn[1][]{#1}
    \maketitle
    \begin{center}
      \vspace{-2pt}
      \includegraphics[width=\textwidth]{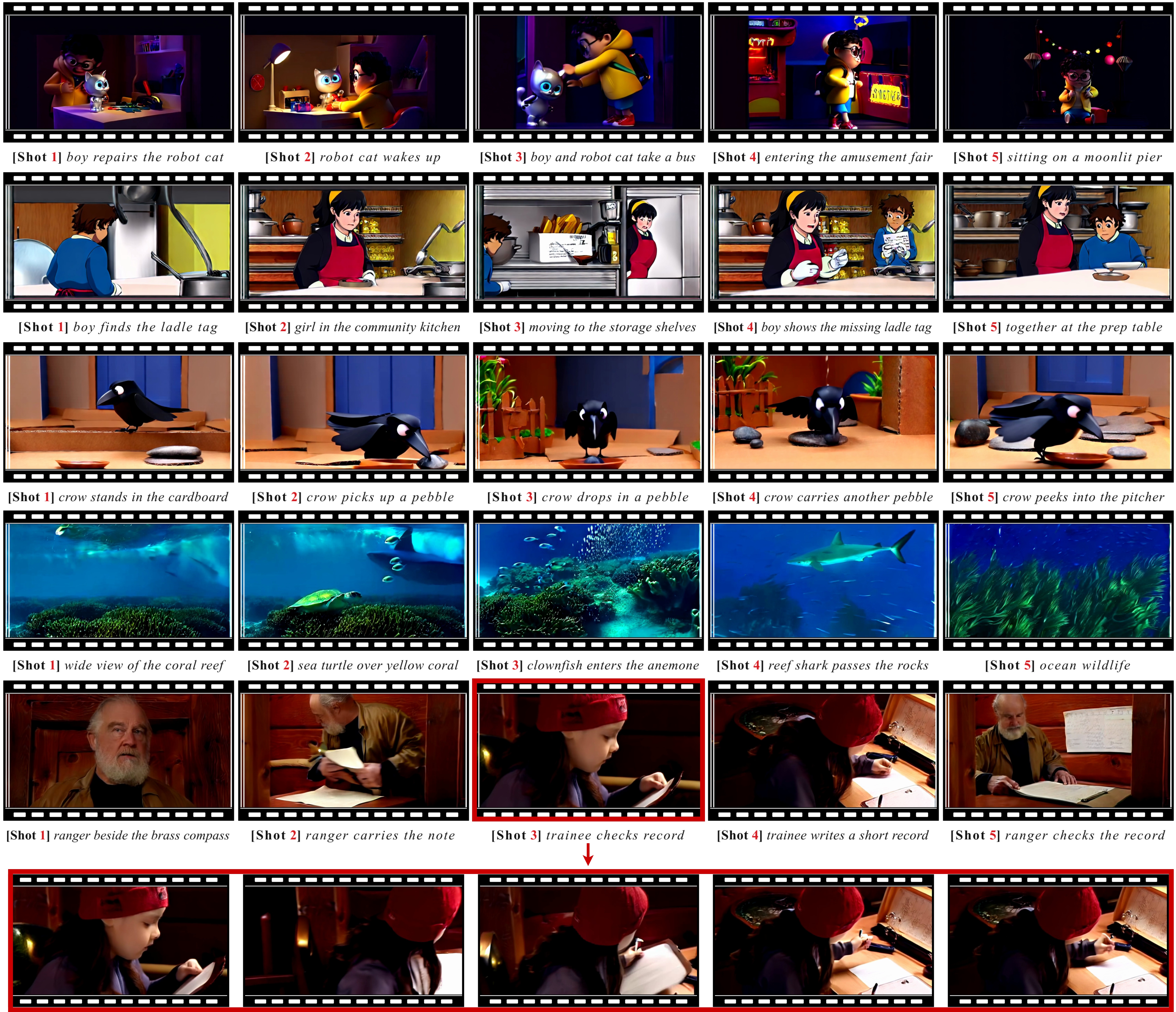}
      \vspace{-10pt}
      \captionsetup{type=figure}
      \caption{\textbf{Long-form multi-shot video generation across diverse visual styles.}
 \name generates coherent multi-shot sequences in 3D CGI (row~\textcolor{red}{1}), anime (row~\textcolor{red}{2}), stop-motion (row~\textcolor{red}{3}), realistic (row~\textcolor{red}{4}), and cinematic (row~\textcolor{red}{5}) styles. Each row shows temporally ordered shots from a single generated video. Despite changes in viewpoint, shot scale, and scene composition, subjects, environments, and overall visual appearance remain consistent across shots. The enlarged sequence (\textcolor{red}{\textbf{red bounding box sequence}}) at the bottom further shows temporally coherent motion within a shot, together with stable continuous motion and consistent fine-grained scene details over time.}
      \label{fig:teaser}
      \vspace{10pt}
    \end{center}
}]

% \maketitle
% \thispagestyle{empty}

\begin{abstract}
Generating long-form multi-shot videos requires temporally coherent motion within each shot and visually consistent transitions across many shots.
However, most existing video generators are biased toward preserving continuous motion over presenting the full shot sets, and packing an entire multi-shot narrative along a single temporal axis (\ie, Temporal Packing) reinforces the bias.
This motivates decomposing a long video generation into producing shorter video chunks, so that each temporal axis handles fewer shots and thus better models continuous motion. 
Since independently generated video chunks cannot directly establish consistent narratives, we arrange them on a spatial grid for joint modeling.
We therefore propose \name, a Multi-Grid Post-Training paradigm for long-form multi-shot video generation. 
To support this paradigm, we construct the Multi-Grid Long Video (MGLV) dataset from 1,000 long-form videos through Source Video Collection, Hierarchical Video Segmentation, Grid Video Construction, and Character-Aware Story Annotation, yielding 54K grid videos paired with video story prompts.
To model the grid structure and support conditional extension across grid videos, our Noise-Free Random-Grid Training retains a random subset of video chunks in the grid as clean visual context to guide the denoising of the remaining video chunks.
Furthermore, we employ the Grid Embedding to encode specific video grid spatial information, the character-aware Story Prompt links shared entities across video chunks, and the Grid Boundary Loss stabilizes the grid structure.
Under the same token budget, our \name generates 6.05$\times$ more video shots than the Temporal Packing baseline in a 1,616-frame video.
Compared with other methods, \name achieves state-of-the-art intra-shot consistency (0.9131 vs. 0.8086 for HoloCine) and inter-shot consistency (0.5914 vs. 0.5384 for StoryMem) on our curated video benchmark spanning 5 real-world categories. 
Further experiments validate that \name can scale the video length with minimal compromise via a single or multiple generations.
\end{abstract}
\begin{figure}[ht]    \centering    \includegraphics[width=\columnwidth]{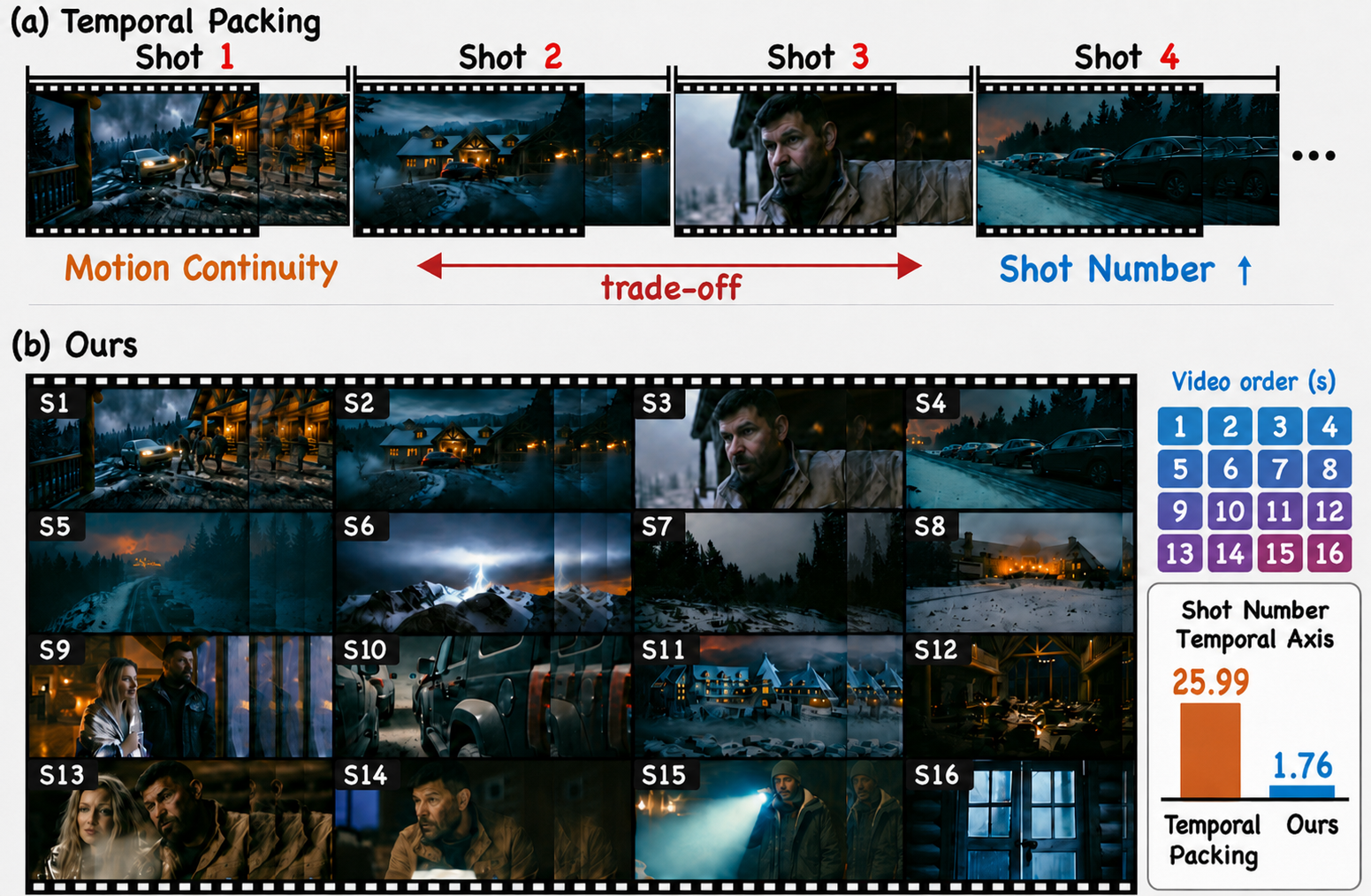}

\caption{\textbf{\name reduces the number of shots per temporal axis through spatial grid modeling.}
(a) The plain Temporal Packing method places an entire multi-shot narrative along a single temporal axis, requiring the model to handle a growing shots and encouraging its bias toward preserving continuous motion over presenting the full set of shots.
(b) \name decomposes the long video narrative into short video chunks and spatially arranges them in a unified grid video for modeling. Each temporal axis handles fewer shots, reducing the average number of shot transitions per temporal axis from \(25.99\) to \(1.76\).}

\label{fig:packing}    
\vspace{-10pt}
\end{figure}

\section{Introduction}\label{sec:intro}

Cinematic narratives rarely unfold in a single continuous shot; instead, they are conveyed through sequences of shots that vary in viewpoint, scale, and scene~\cite{rao2020unified,huang2020movienet}. Such shot-based storytelling imposes two complementary requirements: (i) coherent motion within each shot and (ii) consistency of characters, environments, and narrative progression across shots. These requirements become increasingly difficult to satisfy as videos grow longer and contain more shots, since recurring entities and story states must remain stable across increasingly distant and visually diverse contexts~\cite{zhao2025moviedreamer,he2026entitybench,lai2026groundshot}. Despite remarkable advances in visual quality and motion realism~\cite{kong2024hunyuanvideo,polyak2024movie,yang2025cogvideox,wan2025}, current video generation models still struggle to meet both requirements over long-form multi-shot sequences.

Existing approaches to this long-horizon problem generally fall into three paradigms: autoregressive extension~\cite{zhang2025storymem,luo2026shotstream,an2026onestory}, keyframe-based interpolation~\cite{zhou2024storydiffusion,zhang2026stage,xiao2026captain}, and holistic joint generation~\cite{wang2026multishotmaster,meng2026holocine,kara2025shotadapter,guo2025long}. 
Autoregressive methods~\cite{zhang2025storymem,luo2026shotstream,henschel2025streamingt2v,gao2024vid,xie2025progressive} extend videos sequentially and preserve local temporal continuity, but repeated conditioning on previously generated content makes them prone to error accumulation, while maintaining longer histories incurs increasing memory costs. 
Keyframe- or storyboard-guided methods~\cite{zhou2024storydiffusion,zhang2026stage,xiao2026captain} anchor selected narrative states to improve structural control, but sparse visual anchors do not directly constrain motion and appearance throughout the generated sequence. 
Holistic methods~\cite{wang2026multishotmaster,meng2026holocine,kara2025shotadapter,guo2025long} jointly process all shots to facilitate global coordination. 
However, most video generators are biased toward preserving continuous motion over presenting the full shot sets, and packing an entire multi-shot narrative along a single temporal axis (\ie, Temporal Packing) reinforces this bias.

To overcome this bias, we decompose a long-form multi-shot narrative into multiple temporally ordered short video chunks, distributing the full set of shots across shorter temporal axes so that each axis handles fewer shots and can better model continuous motion. 
Since independently generated video chunks cannot directly establish a consistent narrative, we propose \name, a Multi-Grid Post-Training paradigm that spatially arranges these chunks in a unified grid for joint generation, enabling cross-chunk information exchange and global narrative consistency (\cref{fig:packing}). 
Existing grid-based formulations serve different purposes: Grid Diffusion Models~\cite{lee2024grid} tile individual video frames into a 2D grid image, converting temporal positions into spatial locations for efficient text-to-video generation, whereas VIC~\cite{fei2024video} concatenates video clips spatially or temporally as an in-context interface between observed and target videos for conditional completion.  
In contrast, \name uses a spatial grid as the joint generative representation of consecutive video chunks from a single long video. Each video chunk evolves along a local temporal axis; all video chunks are jointly modeled to coordinate characters, environments, and narrative progression across the grid. Generated video chunks are finally unpacked in temporal order to form a long-form multi-shot video.

\begin{figure*}[t] 
    \centering
    \includegraphics[width=\textwidth]{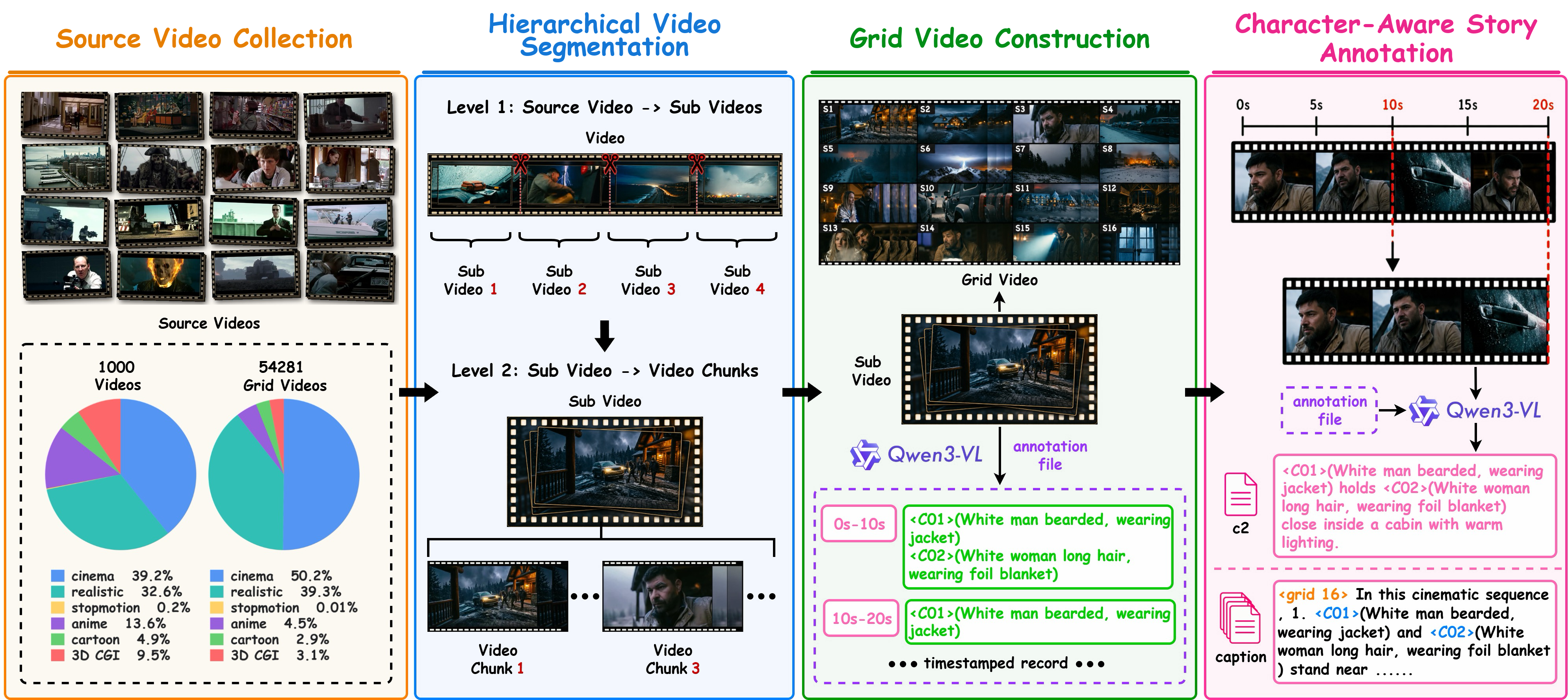} 
\caption{\textbf{Construction and annotation pipeline for MGLV.}
From 1,000 long-form videos, we divide each temporal segment into short video chunks.
Rather than concatenating them along the temporal axis, we spatially arrange the chunks in temporal order to form a grid video. 
Qwen3-VL first extracts timestamped character descriptions and then uses these descriptions to caption the corresponding temporal intervals. These interval-level captions are aggregated into character-aware story prompts prefixed with \texttt{<grid N>}. The resulting MGLV dataset comprises 54,281 grid videos spanning diverse visual styles.}
    \vspace{-2ex} 
    \label{fig:dataset} 
\end{figure*}

To support \name, we construct the Multi-Grid Long Video (MGLV) dataset (\cref{fig:dataset}) from 1,000 long-form source videos through a four-stage pipeline consisting of \textit{Source Video Collection}, \textit{Hierarchical Video Segmentation}, \textit{Grid Video Construction}, and \textit{Character-Aware Story Annotation}. This pipeline yields 54K grid videos, each comprising temporally ordered video chunks and paired with video story prompts.
Building on MGLV, we introduce four complementary components in \name: (1) \textit{Noise-Free Random-Grid Training}, which randomly keeps a subset of video chunks as noise-free visual context to guide the denoising of the remaining chunks, also enabling conditional generation across successive grid videos; 
(2) \textit{Grid Embedding}, which augments each latent token with its grid identity, grid geometry, and intra-grid position to provide spatial and structural cues; 
(3) \textit{Story Prompts}, which link recurring entities across video chunks using shared character tags for character-consistent generation; and 
(4) \textit{Grid Boundary Loss}, which explicitly supervises grid boundaries to stabilize the generated grid structure.

As shown in \cref{fig:teaser}, \name generates coherent long-form multi-shot videos across diverse visual styles, maintaining consistent subjects and environments across shots while preserving continuous motion within each shot. Under the same token budget, \name generates 1,616-frame multi-shot videos with 6.05 times more shots than the plain Temporal Packing baseline (\cref{fig:shot_count}). 
Compared with existing methods, \name achieves state-of-the-art (SoTA) intra-shot consistency for subjects (0.8970 vs. 0.7814 for HoloCine~\cite{meng2026holocine}) and backgrounds (0.9291 vs. 0.8358), as well as inter-shot consistency for subjects (0.6139 vs. 0.5543 for StoryMem~\cite{zhang2025storymem}) and backgrounds (0.5689 vs. 0.5224) (\cref{tab:evaluation} and \cref{fig:consistency}).
\name further scales video length in two complementary ways. Increasing the grid count from 16 to 64 extends the unpacked sequence from 1,616 to 6,464 frames without increasing the video latent tokens processed by the model (\cref{fig:grid_scaling}). To alleviate the trade-off between video length and resolution, \name can further extend generation across successive grid videos by conditioning on previously generated video chunks (\cref{fig:continue}).

\section{Related Work}\label{sec:related}

\subsection{Video Diffusion Models}
Text-to-video generation has evolved from spatiotemporal U-Net diffusion models~\cite{ho2022video,ho2022imagen,singer2022make,blattmann2023align} to large-scale latent video Diffusion Transformers (DiTs)~\cite{ma2024latte,yang2025cogvideox,kong2024hunyuanvideo,wan2025}, substantially improving visual fidelity, temporal dynamics, and prompt alignment.
Recent systems further scale this paradigm through large-scale pretraining, latent compression, and flow-matching objectives~\cite{chen2023videocrafter1,blattmann2023stable,lin2024open,polyak2024movie,hacohen2024ltx}.
Complementary approaches introduce reference images and camera motion as additional conditions for controllable generation~\cite{jiang2024videobooth,yuan2025identity,zhang2025tora,wang2024motionctrl,he2024cameractrl,wu2024draganything}.
Despite these advances, most general video models remain primarily optimized for short clips with a single shot.

\subsection{Long-Form Multi-Shot Video Generation}
Recent studies extend video generation from short clips to long-form multi-shot narratives while preserving recurring characters, environments, and styles~\cite{kara2025shotadapter,zheng2024videogen,meng2026holocine,wang2026multishotmaster,zhang2025storymem,luo2026shotstream}.
Autoregressive approaches generate successive shots by propagating preceding frames, feature caches, or visual memories~\cite{henschel2025streamingt2v,sun2025ar,yin2025slow,zhang2025storymem,luo2026shotstream,xie2025progressive}, but recursive conditioning cause accumulate errors and visual drift.
Storyboard-based methods generate keyframes as visual anchors and then expand them into individual video segments~\cite{zhou2024storydiffusion,zheng2024videogen,zhang2026stage,zhao2025moviedreamer,xiao2026captain}, concentrating consistency constraints mainly at sparse narrative states.
Other methods jointly generate multiple shots with cross-shot attention and shot-aware conditioning~\cite{guo2025long,kara2025shotadapter,wang2026multishotmaster,meng2026holocine}, yet place the full narrative along an extended temporal axis that must model both continuous motion and discrete shot transitions.
In contrast, \name reorganizes temporally ordered video chunks into a spatial grid for joint generation, reducing the number of shot transitions assigned to each local temporal axis.

\subsection{Grid-Structured Visual Generation}
Grid-structured representations have been explored for image generation, visual in-context learning, and image editing~\cite{zeng2024jedi,li2025visualcloze,zhang2026enabling,tomar2025gridit,lee2024grid,fei2024video}.
JeDi~\cite{zeng2024jedi} learns the joint distribution of multiple images sharing a common subject for personalized generation, while VisualCloze~\cite{li2025visualcloze} and ICEdit~\cite{zhang2026enabling} organize visual inputs and outputs on a shared canvas for in-context generation and image editing.
Grid Diffusion Models~\cite{lee2024grid} and GriDiT~\cite{tomar2025gridit} arrange video frames into 2D grids, where each grid region represents a single frame rather than a video segment with an explicit temporal dimension.
VIC~\cite{fei2024video} concatenates video clips spatially or temporally and uses reference clips to condition the generation of target clips.
In contrast, \name represents a long-form multi-shot narrative as an ordered grid of temporally evolving video chunks: each grid region retains a local temporal axis, while all chunks are jointly generated and unpacked in temporal order.
\section{MGLV Dataset}\label{sec:dataset}

To support multi-grid post-training, we construct the Multi-Grid Long Video (MGLV) dataset, comprising 54,281 grid videos derived from 1,000 long-form source videos and paired with character-aware annotations. As illustrated in \cref{fig:dataset}, constructing MGLV involves four stages: source video collection, hierarchical video segmentation, grid video construction, and character-aware story annotation.

\subsection{Source Video Collection}

We collect 1,000 long-form source videos ranging in duration from 3 minutes to 4 hours and spanning cinematic, realistic, anime, cartoon, stop-motion, and 3D CGI visual styles. We retrieve these publicly available YouTube videos and manually verify them to remove low-quality, duplicate, or unsuitable content and trim irrelevant opening and ending segments from the collected videos. The retained videos contain frequent shot transitions and recurring characters, objects, and environments, providing natural supervision for multi-shot learning and cross-shot consistency.

\subsection{Hierarchical Video Segmentation}

After resampling all source videos to 30 FPS, we partition each video into multiple 1,296-frame subvideos, each further divided into 16 non-overlapping 81-frame video chunks. Importantly, the segmentation is based on fixed temporal intervals rather than detected shots---a video chunk does not necessarily correspond to a single shot. 
This detector-free design avoids the overhead and errors of shot-boundary detection, enabling scalable dataset construction.

\begin{figure*}[t] 
    \centering
    \includegraphics[width=\textwidth]{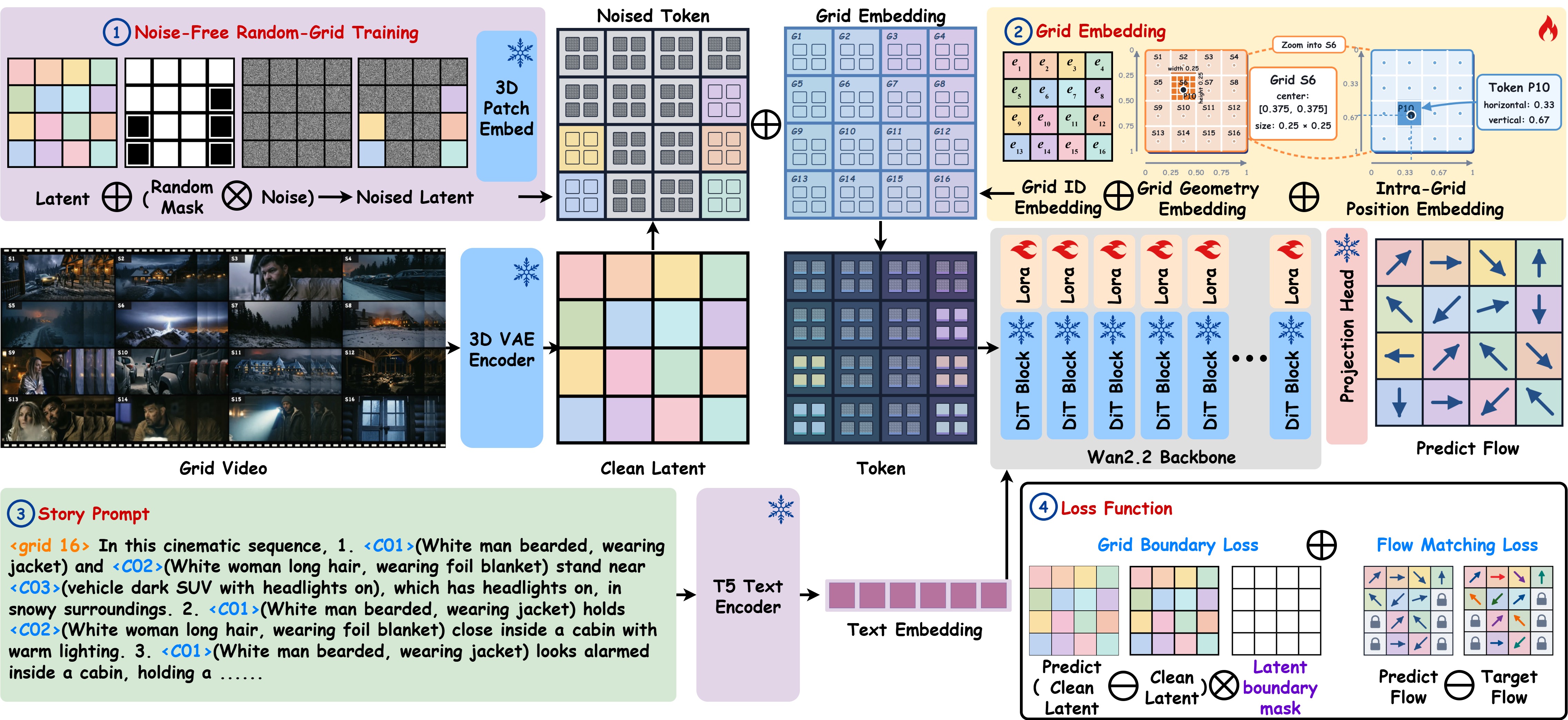} 
    \caption{\textbf{Overview of \name.} It generates all temporally ordered video chunks packed into a grid video. \textit{Noise-Free Random-Grid Training} preserves selected grids as clean visual context, while \textit{Grid Embedding} and \textit{Story Prompts} provide structural and semantic conditioning. Training combines the standard Flow Matching Loss with \textit{Grid Boundary Loss} to preserve the grid structure.} 
    \vspace{-2ex} 
    \label{fig:overview} 
\end{figure*}

\subsection{Grid Video Construction}

For each subvideo, we arrange its 16 video chunks in the chronological order on a $4\times4$ spatial grid. The resulting 81-frame grid video represents all 1,296 original frames, with their temporal order encoded by the fixed ordering of the grids. 
We further quantify the shot density of MGLV using TransNetV2~\cite{soucek2024transnet}. Each subvideo contains \(25.99\) shots on average, whereas each video chunk contains only \(1.76\) shots (\cref{fig:packing}). Despite using fixed video segmentation without constraining the shot boundary, \name reduces the average shot load along each modeled chunk by \(14.8\times\).

\subsection{Character-Aware Story Annotation}

To provide temporally grounded supervision for recurring entities, we use Qwen3-VL 8B~\cite{Qwen3-VL} in a two-stage pipeline to annotate each grid video. In the first stage, Qwen3-VL analyzes the full subvideo to produce timestamped character records, each linking a temporal interval to descriptions of the characters appearing within it. In the second stage, Qwen3-VL conditions on these records to generate captions for the corresponding temporal intervals, grounding each event description in the characters present at that time. Finally, we concatenate the interval-level captions in temporal order to form the \textit{Story Prompt}. 
We augment it with two special tokens: a leading \texttt{<grid N>} declares the grid configuration containing \(N\) video chunks, while \texttt{<C>}s link recurring entities across video chunks.

\section{Method}\label{sec:method} 

\subsection{Overview}

Let \(V \in \mathbb{R}^{L \times H \times W \times 3}\) denote a video containing \(L\) frames at spatial resolution \(H \times W\). 
Let \(N = R \times C\) be the number of grids. 
Assuming \(L = N \times T\), we partition \(V\) along the temporal axis into \(N\) ordered video chunks \(\{S_i\}_{i=1}^{N}\), where each \(S_i \in \mathbb{R}^{T \times H \times W \times 3}\) contains \(T\) consecutive frames.
% At each local time step \(\tau \in \{1,\ldots,T\}\), 
We spatially tile the corresponding frames from all \(N\) video chunks according to their assigned grids, producing a grid video \(X \in \mathbb{R}^{T \times (R \times H) \times (C \times W) \times 3}\). \name therefore transfers a factor of \(N\) from the temporal dimension to the spatial grid, representing all \(N \times T\) frames using only \(T\) temporal steps without discarding any frames.
At inference, we spatially unpack the generated grid video \(\hat{X}\) into grid-wise video chunks \(\{\hat{S}_i\}_{i=1}^{N}\) and concatenate them along the temporal axis in ascending grid-index order to obtain the video \(\hat{V} \in \mathbb{R}^{(N \times T) \times H \times W \times 3}\).
Given a grid video and its story prompt $c$, \name uses Noise-Free Random-Grid Training that randomly selects video chunks as clean visual context while noising the remaining chunks for joint denoising, and augments grid spatial information with grid embedding. 
As illustrated in \cref{fig:overview}, \name jointly denoises the grid video latents conditioned on the story prompt, with flow matching loss and grid boundary loss supervising video generation and grid structure, respectively.

\subsection{Multi-Grid Post-Training} 
 
\paragraph{Noise-Free Random-Grid Training.} In each grid video, \name keeps a randomly selected subset of video chunks noise-free during training while applying the standard diffusion noising process to the remaining video chunks. 
Specifically, for each training sample, we activate noise-free random grid conditioning with probability $p_{\mathrm{vis}}=0.3$. 
When activated, we sample $N_{\mathrm{vis}}\sim\mathcal{U}\{1,\ldots,8\}$ and uniformly select $N_{\mathrm{vis}}$ distinct video chunks in the grid to remain noise-free. Let $\mathbf{M}_{\mathrm{vis}}$ denote the grid-wise binary mask marking the selected noise-free grids, broadcast to the corresponding grid video latent positions. The forward process is defined as:
\begin{equation}\begin{aligned}\mathbf{z}_1 &= \mathcal{E}_{\mathrm{VAE}}(\mathbf{X}), \qquad \boldsymbol{\epsilon} \sim \mathcal{N}(\mathbf{0},\mathbf{I}), \\\mathbf{z}_t &= (1-t)\mathbf{z}_1 + t\boldsymbol{\epsilon}, \\\tilde{\mathbf{z}}_t &= \mathbf{M}_{\mathrm{vis}} \odot \mathbf{z}_1 + (1-\mathbf{M}_{\mathrm{vis}}) \odot \mathbf{z}_t, \\\mathbf{h}_{0,j} &= \operatorname{PatchEmbed}(\tilde{\mathbf{z}}_t)_j + \mathbf{e}_j^{\mathrm{grid}}, \\\mathbf{H}_0 &= \operatorname{Stack}_{j}(\mathbf{h}_{0,j}), \\\hat{\mathbf{u}}_t &= f_{\theta+\Delta\theta}(\mathbf{H}_0,t,\mathbf{c}), \\\hat{\mathbf{u}}_t^{\mathrm{pred}} &= \operatorname{Select}(\hat{\mathbf{u}}_t,1-\mathbf{M}_{\mathrm{vis}}),\end{aligned}\label{eq:random_grid_forward}\end{equation}
Here, $\mathcal{E}_{\mathrm{VAE}}$ is the frozen 3D VAE encoder, $\epsilon$ is standard gaussian noise, and $t$ is the flow-matching timestep. 
The variable $h_{0,j}$ denotes the $j$-th grid video latent token with grid position awareness, formed by adding its token-wise grid embedding $e_j^{\mathrm{grid}}$ to the corresponding grid video latent token, while $H_0$ stacks these tokens to form the model input. 
The conditioning variable $c$ denotes the character-aware story prompt constructed using the annotation pipeline in \cref{sec:dataset}, whereas $\theta$ and $\Delta\theta$ denote the backbone and LoRA parameters~\cite{hu2022lora}, respectively. 
During post-training, the backbone remains frozen, while the LoRA adapters and grid embedding modules are jointly optimized. 
The model first predicts the full flow field $\hat{u}_t$, after which $\operatorname{Select}$ retains only the output flow fields corresponding to the noised grids to obtain $\hat{u}_t^{\mathrm{pred}}$. 
At inference, selected grids from a previously generated grid video can serve as noise-free visual conditions for generating the remaining grids of a new grid video, enabling further extension across successive grid videos without additional training.

\paragraph{Grid Embedding.} To explicitly encode the spatial grid structure introduced by \name, we augment each token with a grid embedding comprising three complementary components: grid identity, grid geometry, and intra-grid position. For the $j$-th token, let $i_j \in \{1,\ldots,N\}$ denote its associated grid index. The corresponding token-wise grid embedding is defined as:
\begin{equation}
\mathbf{e}_j^{\mathrm{grid}} = \mathbf{e}_{i_j}^{\mathrm{id}} + \mathbf{E}_{\mathrm{geo}}(\boldsymbol{\gamma}_{i_j}) + \mathbf{E}_{\mathrm{pos}}(\mathbf{r}_j).\label{eq:grid_embedding}
\end{equation} 
Here, $e_{i_j}^{\mathrm{id}}$ is a learned \textit{Grid ID Embedding}. The geometry vector
$\gamma_{i_j}=(c_{i_j}^{x},c_{i_j}^{y},w_{i_j},h_{i_j})$
encodes the center coordinates and spatial dimensions (width and height) of grid $i_j$, all expressed in normalized units, while
$r_j=(u_j,v_j)$
represents the normalized position of the token within that grid. 
Both $E_{\mathrm{geo}}$ and $E_{\mathrm{pos}}$ are implemented as two-layer MLPs with SiLU activations, each projecting its input to the transformer hidden dimension to obtain \textit{Grid Geometry Embedding} and \textit{Intra-Grid Position Embedding}. Together, these components distinguish tokens associated with different grids while encoding their grid-local positions in a shared normalized coordinate system.

\begin{table*}[tb]
  \scriptsize
  \caption{\textbf{Quantitative results on our curated video benchmark}. Intra- and inter-shot consistency measure visual scene preservation within and across shots, while aesthetic quality, dynamic degree, and semantic alignment assess visual quality, motion magnitude, and prompt fidelity. We mark the \textbf{best} and the \underline{second best} results.}
  \label{tab:evaluation}
  \centering
  \resizebox{\textwidth}{!}{%
    \renewcommand{\arraystretch}{1.2}%    \setlength{\tabcolsep}{4pt}%
    \begin{tabular}{p{3.4cm}p{2.0cm}ccccccc}
      \toprule
      \multirow{2}{*}{\textbf{Method}}
      & \multirow{2}{*}{\textbf{Backbone}}
      & \multicolumn{2}{c}{\textbf{Intra-Shot Consistency}}
      & \multicolumn{2}{c}{\textbf{Inter-Shot Consistency}}
      & \multirow{2}{*}{\parbox{1.5cm}{\centering
          \textbf{Aesthetic Quality}\textuparrow}}
      & \multirow{2}{*}{\parbox{1.5cm}{\centering
          \textbf{Dynamic Degree}\textuparrow}}
      & \multirow{2}{*}{\parbox{1.5cm}{\centering
          \textbf{Semantic Align.}\textuparrow}} \\
      \cline{3-4}\cline{5-6}
      & & \textbf{Subject}\textuparrow
      & \textbf{Background}\textuparrow
      & \textbf{Subject}\textuparrow
      & \textbf{Background}\textuparrow
      & & & \\
      \hline

      Wan2.2+StoryDiffusion~\cite{zhou2024storydiffusion}
      & Wan2.2 5B
      & 0.3376
      & 0.3658
      & 0.2707
      & 0.2684
      & 0.2572
      & 0.3396
      & 0.1014 \\ 
      
      Wan2.2+STAGE~\cite{zhang2026stage}
      & Wan2.2 5B
      & 0.5982
      & 0.6501
      & 0.3717
      & 0.3538
      & 0.4032
      & 0.5701
      & 0.1771 \\ \cmidrule(lr){1-9}

      Mask\textsuperscript{2}DiT~\cite{qi2025mask}
      & CogVideoX 5B
      & 0.5701
      & 0.5772
      & 0.2649
      & 0.3344
      & 0.3431
      & 0.0787
      & 0.1266 \\

      ShotStream~\cite{luo2026shotstream}
      & Wan2.1 1.3B
      & 0.6110
      & 0.6281
      & 0.4142
      & 0.3691
      & 0.3887
      & 0.1366
      & 0.1454 \\ 
      
      StoryMem~\cite{zhang2025storymem}
      & Wan2.2 14B
      & 0.7295
      & 0.7688
      & \underline{0.5543}
      & \underline{0.5224}
      & 0.4899
      & \textbf{0.7573}
      & \underline{0.2060} \\ \cmidrule(lr){1-9}

      Wan2.2~\cite{wan2025}
      & Wan2.2 5B
      & 0.7157
      & 0.7465
      & 0.5181
      & 0.3766
      & 0.4959
      & 0.4388
      & 0.2034 \\

      \mbox{Wan2.2 Temporal Packing~\cite{wan2025}}
      & Wan2.2 5B
      & 0.2018
      & 0.2067
      & 0.0564
      & 0.0589
      & 0.1019
      & 0.1589
      & 0.0416 \\ 
      MultiShotMaster~\cite{wang2026multishotmaster}
      & Wan2.1 1.3B
      & 0.6969
      & 0.7186
      & 0.4001
      & 0.3213
      & 0.4319
      & 0.4342
      & 0.1780 \\
      
      HoloCine~\cite{meng2026holocine}
      & Wan2.2 14B
      & \underline{0.7814}
      & \underline{0.8358}
      & 0.5187
      & 0.4287
      & \textbf{0.5151}
      & 0.6916
      & \textbf{0.2147} \\ \cmidrule(lr){1-9}

      VIC~\cite{fei2024video}
      & HunyuanVideo 13B
      & 0.0268
      & 0.0302
      & 0.0152
      & 0.0399
      & 0.0128
      & 0.0146
      & 0.0058 \\

      Wan2.2+VIC-Style~\cite{fei2024video}
      & Wan2.2 5B
      & 0.3144
      & 0.3580
      & 0.1571
      & 0.2806
      & 0.2523
      & 0.3228
      & 0.1015 \\

    \rowcolor[HTML]{EAF4FB}
    \name (Ours)
    & Wan2.2 5B
    & \textbf{0.8970}
    & \textbf{0.9291}
    & \textbf{0.6139}
    & \textbf{0.5689}
    & \underline{0.5087}
    & \underline{0.7420}
    & 0.1959 \\

      \bottomrule
    \end{tabular}%
  }
\end{table*}

\paragraph{Loss Function.} We optimize \name using a joint objective that combines the standard flow-matching loss $\mathcal{L}_{\mathrm{FM}}$, evaluated on flow fields from the noised grids, with a grid boundary loss $\mathcal{L}_{\mathrm{GB}}$ that focuses reconstruction supervision on the latent spatial boundaries between adjacent grids. 
The overall objective is as follows:
\begin{equation}
\begin{aligned}
\mathcal{L}&= \mathcal{L}_{\mathrm{FM}}\!\left(\hat{\mathbf{u}}_{t}^{\mathrm{pred}}, \operatorname{Select}\!\left(\boldsymbol{\epsilon}-\mathbf{z}_{1}, 1-\mathbf{M}_{\mathrm{vis}}\right)\right) \\&\quad + \lambda_{\mathrm{GB}}\mathcal{L}_{\mathrm{GB}}.
\end{aligned}
\label{eq:overall_loss}
\end{equation} 
where $\lambda_{\mathrm{GB}}=0.1$ controls the relative contribution of the grid boundary loss.
To compute $\mathcal{L}_{\mathrm{GB}}$, we first estimate the clean latents from the noised latents $\tilde{z}_t$ and the full predicted flow fields $\hat{u}_t$, and then evaluate their discrepancy from the ground-truth clean latents $z_1$ only at grid-boundary locations. Specifically, using the known grid layout, we construct a fixed binary grid-boundary mask $\mathbf{B}$ in latent coordinates, assigning ones to grid-boundary locations and zeros to the interior of each grid. $\mathcal{L}_{\mathrm{GB}}$ is then defined as
\begin{equation}
\begin{aligned}
\hat{\mathbf{z}}_{1}
&= \tilde{\mathbf{z}}_{t}-t\hat{\mathbf{u}}_{t}, \\
\mathcal{L}_{\mathrm{GB}}
&=
\frac{
\left\|
\mathbf{B}\odot
\left(\hat{\mathbf{z}}_{1}-\mathbf{z}_{1}\right)
\right\|_{2}^{2}
}{
\|\mathbf{B}\|_{1}
}.
\end{aligned}
\label{eq:grid_boundary_loss}
\end{equation}

\begin{figure*}[t] 
    \centering
    \includegraphics[width=1\textwidth]{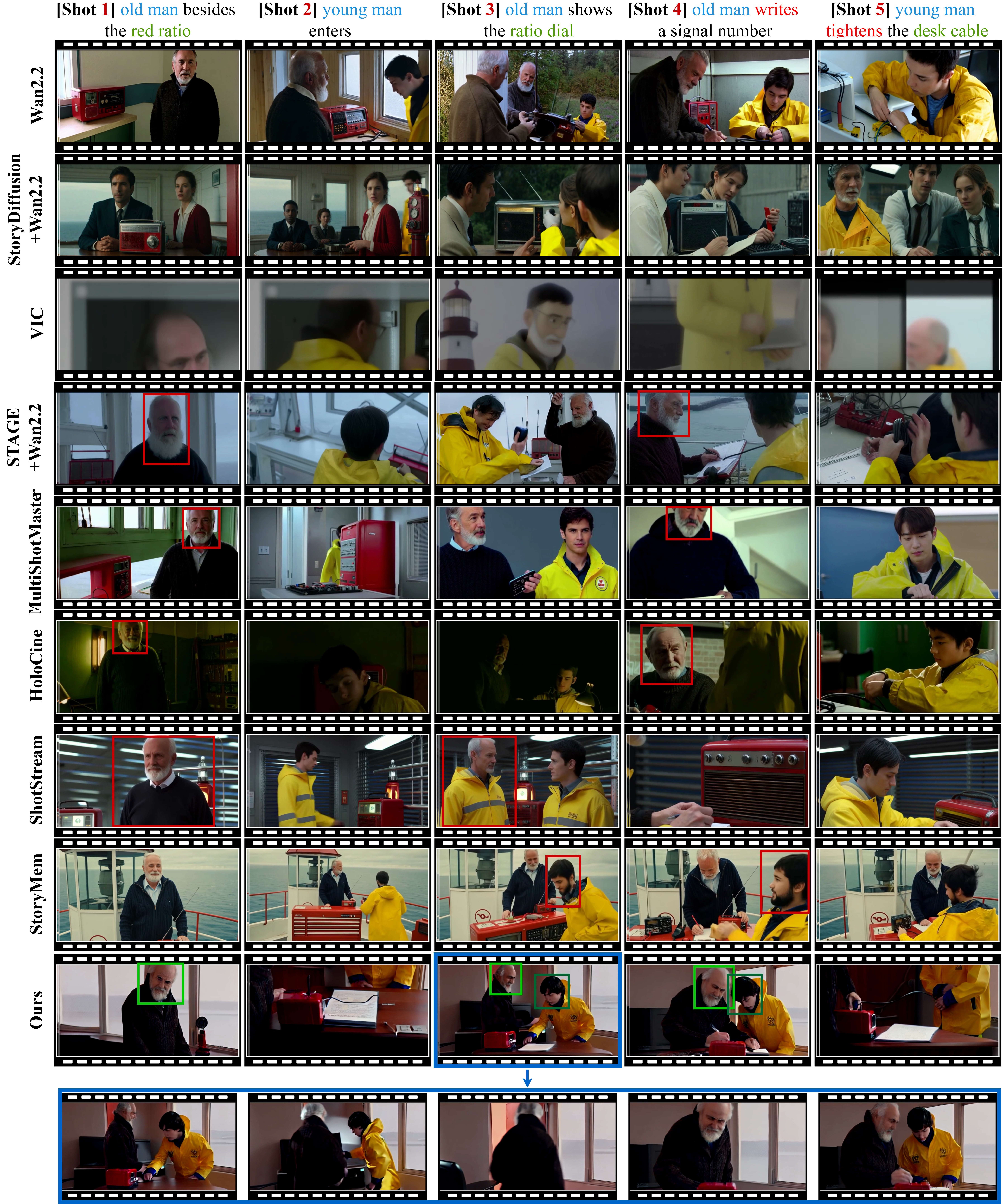} 
    \caption{\textbf{Qualitative comparison.}
Baselines exhibit identity drift, whereas \name preserves recurring characters and scenes across shot transitions. \textbf{Bounding boxes} highlight identity drift and preservation; the enlarged sequence shows coherent within-shot motion.}
    % \vspace{-2ex} 
    \label{fig:comparison} 
\end{figure*}

\begin{figure*}[t] 
    \centering
    \includegraphics[width=\textwidth]{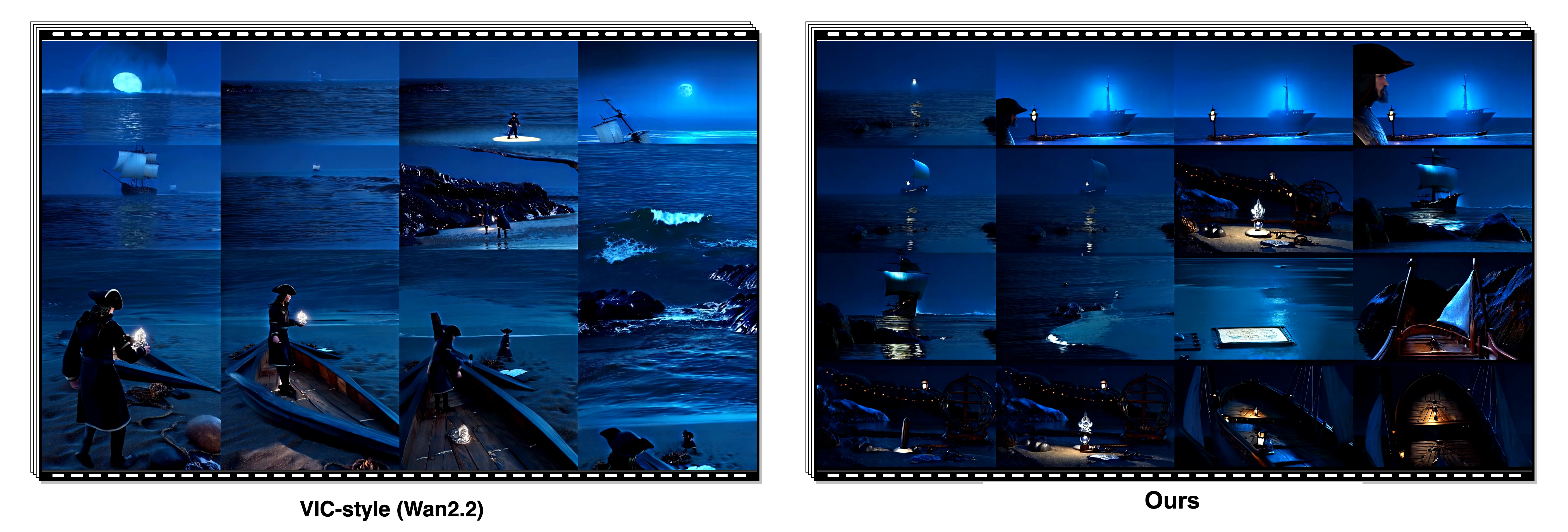} 
\caption{\textbf{Qualitative comparison with VIC-style training.}
Under matched training settings, VIC-style training fails to preserve the complete grid structure, whereas \name produces a well-separated \(4\times4\) grid of coherent video chunks.} 
    \vspace{-2ex} 
    \label{fig:vic} 
\end{figure*}

This boundary-focused supervision encourages stable separation between adjacent grids without adding additional constraints on the visual content within each grid.

\section{Experiments}\label{sec:exp}

In this section, we compare \name with representative methods for long-form multi-shot video generation and systematically examine its key design choices. 
% \cref{ssec:implementation} details the experimental setup, and \cref{ssec:comparison} reports quantitative and qualitative comparisons. 
% \cref{ssec:analysis} analyzes the scaling behavior of \name as video length increases. 
% \cref{ssec:ablation} analyzes the effect of packing topology and evaluates the contribution of individual training components. 
% \cref{ssec:case} further evaluates long-range consistency.  

\subsection{Experiment Setup}
\label{ssec:implementation}
\paragraph{Training Setup.}

We build \name on the Wan2.2-5B~\cite{wan2025} backbone and perform post-training on MGLV. Specifically, we train \name on 81-frame grid videos with a fixed canvas resolution of $2560\times1536$ in 10 epochs. Each grid video uses a $4\times4$ layout, yielding a per-grid resolution of $640\times384$. Training runs on 8 NVIDIA B200 GPUs using AdamW, with a global batch size of 8 and a learning rate of $2\times10^{-5}$. We use 100 warmup steps followed by cosine learning-rate decay. We employ rank-32 LoRA adapters, which, together with the Grid Embedding modules, yield a total of 63.1M trainable parameters. 

\paragraph{Baselines.}

We compare \name with representative methods for long-form multi-shot video generation, covering three paradigms: autoregressive extension~\cite{zhang2025storymem,luo2026shotstream}, keyframe interpolation~\cite{zhou2024storydiffusion,zhang2026stage}, and holistic generation~\cite{wang2026multishotmaster,meng2026holocine}. We also include Wan2.2~\cite{wan2025}, Mask$^2$DiT~\cite{qi2025mask}, and VIC~\cite{fei2024video}. For controlled comparisons, we construct two additional Wan2.2-5B baselines---Temporal Packing and VIC-style training---using the same LoRA configuration, training data, and total token budget as \name. Unless otherwise specified, we follow the official inference settings for each baseline to generate 1,616-frame videos, and resize all generated outputs to match \name's output resolution before evaluation.

\paragraph{Evaluation Protocols.}

Our evaluation benchmark comprises 89 diverse stories composed by GPT-5.6-Sol, with no narrative overlap with the training set; each story specifies multiple events across five visual categories: 3D CGI (20), anime (18), stop-motion (12), realistic (19), and cinematic (20) (more details can be found in Appendix~\ref{bmk}).  
Following VBench~\cite{huang2024vbench}, as same as \citet{meng2026holocine,an2026onestory,zhang2025storymem,luo2026shotstream,zhang2026stage}, we measure intra-shot subject and background consistency with DINO~\cite{caron2021emerging} and CLIP~\cite{radford2021learning}, respectively; aesthetic quality with the LAION aesthetic predictor~\cite{schuhmann2022laion}; dynamic degree with RAFT~\cite{teed2020raft}; and semantic alignment with ViCLIP~\cite{wang2024internvid}. For inter-shot consistency, we use Grounding DINO~\cite{liu2024grounding} and SAM~\cite{kirillov2023segment} to localize and segment characters and environments in prompts, and DINOv2~\cite{oquab2023dinov2} to measure the similarity between the same masked regions across shots.

\subsection{Main Results}
\label{ssec:comparison}

\paragraph{MovieGrid Outperforms Baselines.}
\cref{tab:evaluation} shows that \name achieves SoTA intra- and inter-shot consistency for both subjects and backgrounds. For intra-shot subject and background consistency, \name scores \(0.8970\) and \(0.9291\), respectively, outperforming HoloCine (\(0.7814\) and \(0.8358\)). For inter-shot subject and background consistency, it scores \(0.6139\) and \(0.5689\), respectively, outperforming StoryMem (\(0.5543\) and \(0.5224\)). It remains competitive in aesthetic quality, dynamic degree, and semantic alignment. \cref{fig:comparison} further shows that competing methods often exhibit identity drift, inconsistent environments, or incomplete realization of requested shots, whereas \name preserves recurring character identities and scene attributes across viewpoints and compositions, producing coherent multi-shot sequences.

VIC~\cite{fei2024video} targets general video in-context generation with a grid layout resembles to ours, which also spatially concatenates multiple dynamic video chunks for joint modeling. It serves as a comparison to probe whether the gains of \name arise merely from spatial concatenation. 
We include the original VIC and a controlled VIC-style Wan2.2 baseline that extends the layout to 16 grids under matched LoRA and training settings, which removes differences in backbone, grid scale, and training configuration to be a fair comparison. \Cref{fig:comparison} shows degraded outputs from the original VIC, while \cref{fig:vic} shows incomplete grid structure from the controlled baseline; \name instead forms a complete, well-separated grid of coherent video chunks. Relative to this baseline, \name improves intra-shot consistency from \(0.3362\) to \(0.9131\) and inter-shot consistency from \(0.2189\) to \(0.5914\) (\cref{tab:evaluation}). These results indicate that extending VIC-style training to 16 grids alone is less competent than \name \wrt grid integrity and consistency, validating the use of our proposed Multi-Grid Post-Training paradigm.

\begin{figure}[ht]
    \centering
    \includegraphics[width=\columnwidth]{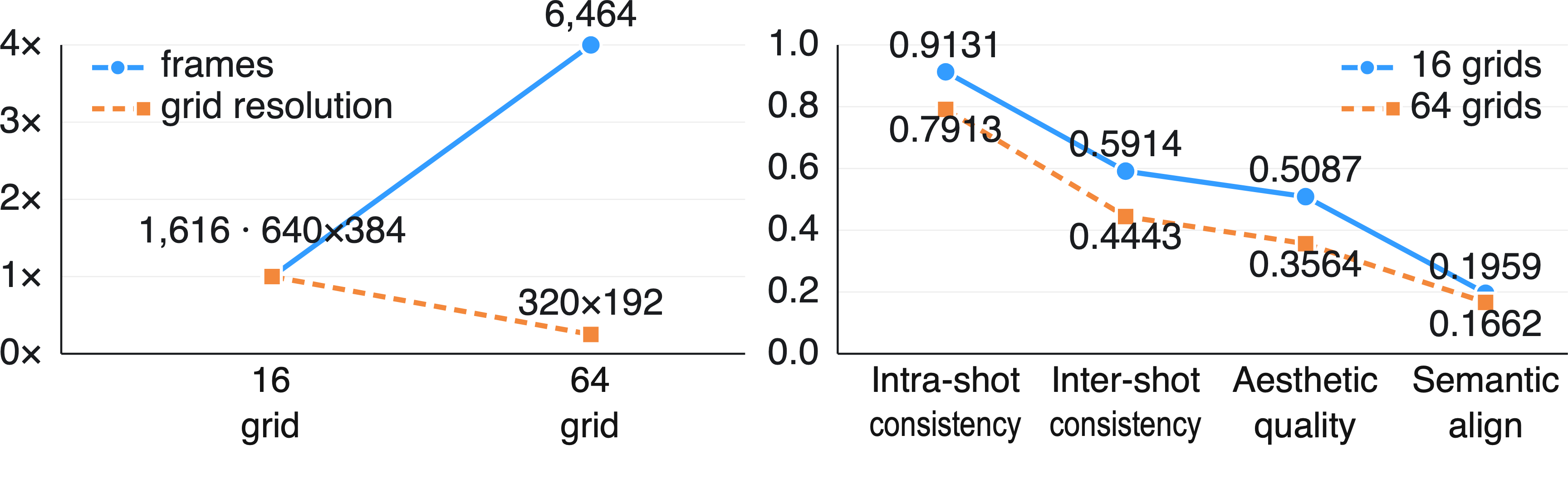}
    \caption{\textbf{Scaling from 16 to 64 grids.} Increasing the grid count quadruples the unpacked video length under a fixed token budget, revealing the resulting duration--resolution--quality trade-off.}
    \label{fig:grid_scaling}
    \vspace{-10pt}
\end{figure}

\subsection{Scaling Video Length}
\label{ssec:analysis}

\paragraph{MovieGrid Scales Length within a Single Generation.} 

Existing methods, though competent in short-video generation, lacks the flexibility to scale to extremely long videos at once due to conventional end-to-end modeling. 
In contrast, \name decouples the story narrative from a single timeline by scaling the parallel video grids, thereby enabling substantially longer multi-shot videos production without increasing the training budget.
To validate this use, we train \name from the original 16-grid (1616 frames) to a 64-grid variant (\name-64) separately on a 64-grid counterpart of MGLV constructed using the same pipeline. 
The corresponding collection contains 16,027 paired 64-grid videos and story prompts. 
\name and \name-64 use \(4\times4\) and \(8\times8\) layouts, respectively, on a fixed \(2560\times1536\) canvas. 
After unpacking, \name-64 yield 6,464 frames videos at \(320\times192\) resolution. 
In \cref{fig:grid_scaling}, this formulation provides a video duration-resolution trade-off: it supports producing longer videos, at the cost of a reasonable compromise in visual consistency and semantic alignment, \eg, 13.45\% drop across intra/inter consistency (75.23\% to 61.78\%) and only a 2.97\% drop (to 16.62\%) in semantic alignment.

\begin{figure}[ht]
    \centering
    \includegraphics[width=\columnwidth]{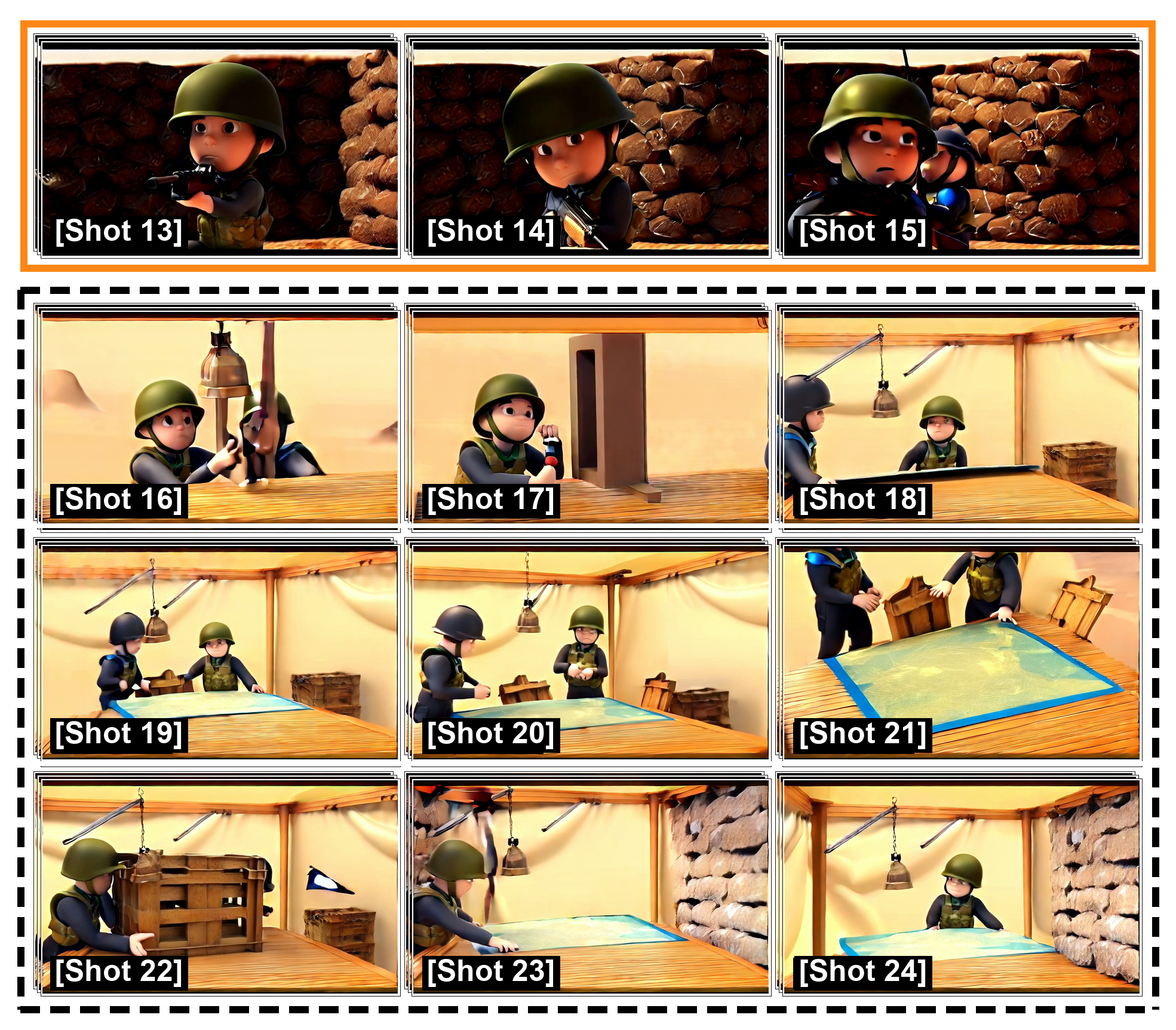}
    \caption{\textbf{Multi-shot story continuation.} \name can generate multi-shot continuation (\textbf{dashed box}) conditioned on existing grid video (\textbf{\textcolor{orange}{orange box}}) with consistent characters.}
    \label{fig:continue}
    \vspace{-10pt}
\end{figure}

\begin{table}[tb]
\small
\caption{\textbf{Ablations on \name components}. Each proposed component contributes to the final \textbf{best results}.}
\label{tab:ablation}
\centering
\resizebox{\linewidth}{!}{%
\setlength{\tabcolsep}{2.5pt} % reduce horizontal column spacing
\renewcommand{\arraystretch}{1.2}%
\begin{tabular}{@{}lcccc@{}}
\toprule

\multirow{2}{*}{\textbf{Method}}
& \multirow{2}{*}{\parbox{1.7cm}{\centering \textbf{Intra-shot.}\textuparrow}}
& \multirow{2}{*}{\parbox{1.7cm}{\centering \textbf{Inter-shot.}\textuparrow}}
& \multirow{2}{*}{\parbox{1.35cm}{\centering \textbf{Aesthetic Quality}\textuparrow}}
& \multirow{2}{*}{\parbox{1.5cm}{\centering \textbf{Semantic.}\textuparrow}} \\
\\

\hline
w/o noise-free random-grid & 0.4720 & 0.4155 & 0.3173 & 0.1186 \\
w/o character tags         & 0.8583 & 0.5719 & 0.4881 & 0.1709 \\
w/o grid embedding         & 0.7250 & 0.3988 & 0.4163 & 0.1385 \\
w/o grid boundary loss     & 0.8412 & 0.4479 & 0.4140 & 0.1370 \\
\name & \textbf{0.9131} & \textbf{0.5914} & \textbf{0.5087} & \textbf{0.1959} \\

\bottomrule
\end{tabular}%
}
\end{table}

\paragraph{MovieGrid Scales Length across Multiple Generations.}
\label{ssec:story_continue}

Scaling from 16 to 64 grids extends video duration at the cost of spatial resolution. To alleviate this trade-off, \name reuses selected video chunks from a preceding grid video as noise-free visual context to condition the next grid video generation, enabling further extension without increasing the grid count or additional training. As shown in \cref{fig:continue}, the continuation introduces a new story while preserving recurring character identities across successive grid videos.

\subsection{Ablation Study}
\label{ssec:ablation}

\begin{figure}[t]
    \centering
    \includegraphics[width=\columnwidth]{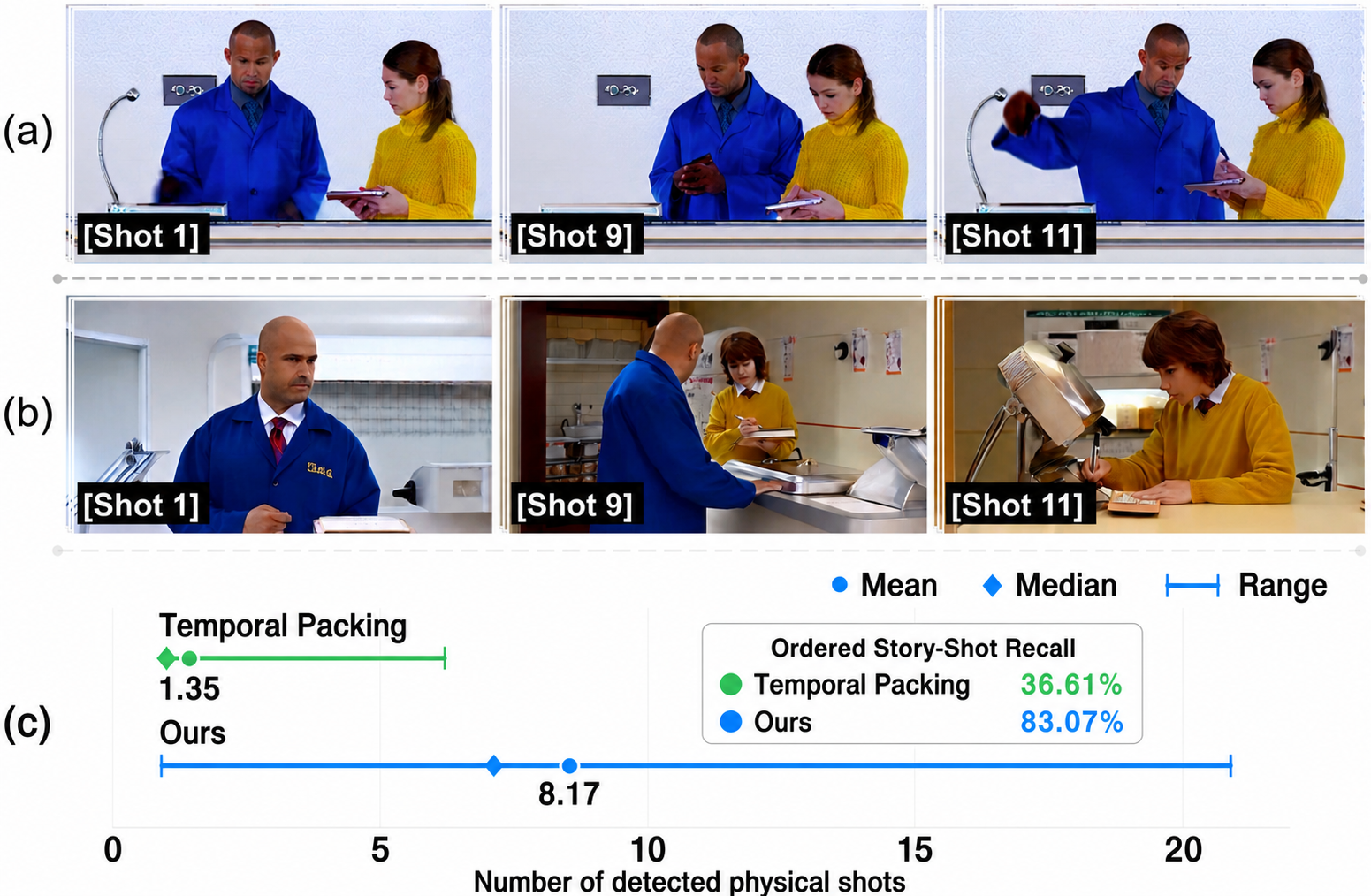}
    \caption{\textbf{\name v.s Temporal Packing.}
    (a) Temporal Packing tends to produce fewer shots, whereas (b) \name realizes more video shots under the same time budget. (c) Shot-count and Ordered Story-Shot Recall show that \name produces more video shots with better story narrative coverage.}
    \label{fig:shot_count}
    \vspace{-10pt}
\end{figure}

\paragraph{Grid Packing Strategy.}
To validate the effect of the proposed packing strategy, we compare \name with the straightforward Temporal Packing baseline under the same budget. 
Temporal Packing assigns the video chunks in longer temporal axis, while \name distributes the same video chunks across spatial grid, shortening each grid’s temporal span. 
To detect shots between different packing strategy, we introduce TransNetV2~\cite{soucek2024transnet} with threshold of \(0.5\). 
Besides, we define Ordered Story-Shot Recall to measures the proportion of shots realized in the
correct temporal order story prompt.
As shown in \cref{fig:shot_count}, TransNetV2 detects an average of \(1.35\) shots for the Temporal Packing baseline (\ie, 78/89 outputs containing only one shot), whereas \(8.17\) shots detected for our \name. 
Ordered Story-Shot Recall increases from \(36.61\%\) to \(83.07\%\), indicating that the additional shots produced align with the video narrative. The paired examples in \cref{fig:shot_count} further illustrate that the straightforward Temporal Packing is increasingly biased towards continuous motion as the target number of shots increases, while \name solve this issue by reducing the expected number of shots modeled along each grid’s temporal axis.

\paragraph{\name Component Ablations.}
\label{ssec:design}
We assess the contribution of different components in \name (\ie, Noise-Free Random-Grid Training, character tags in story prompts, Grid Embedding, Grid Boundary Loss). As shown in \cref{tab:ablation}, removing Noise-Free Random-Grid Training causes the largest degradation---an average drop of 22.14\% across the four metrics (\ie, from 55.23\% to 33.09\%), which validates the effective use of our training paradigm.
Grid Embedding and Grid Boundary Loss also lead to substantial drops of average 13.26\% and 9.23\% (\ie, 55.23\% v.s 41.97\% and 46.00\%), respectively, while character tags have a smaller but consistent effect (3.00\% drop to 52.23\%). Overall, these results indicate that each component contributes consistently to the final performance.

\subsection{Case Study}
\label{ssec:case}

\begin{figure}[t]
    \centering
    \includegraphics[width=\columnwidth]{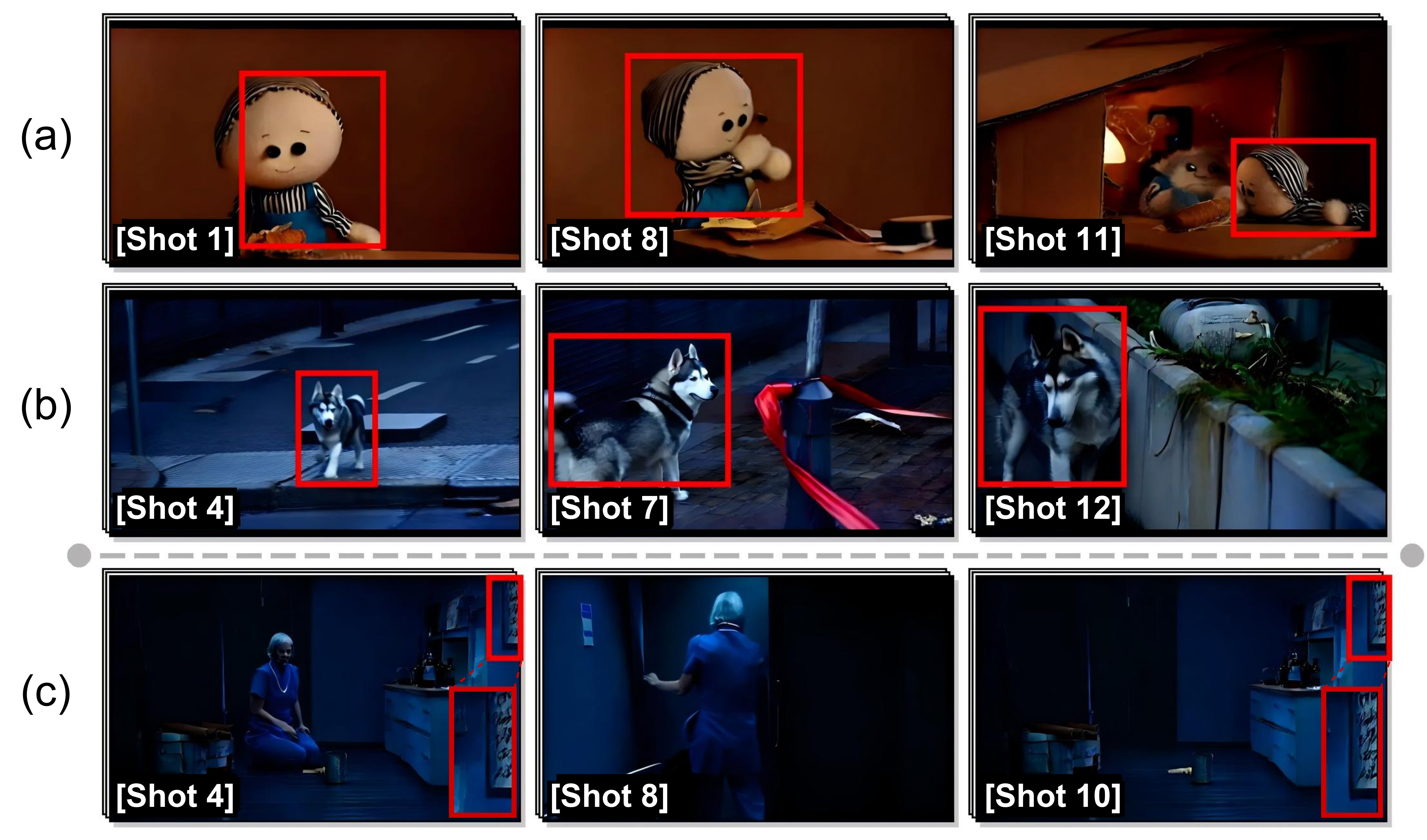}
    \caption{\textbf{Long-range visual consistency.} \name preserves (a) character identity and appearance across changes in shot, (b) non-human subject identity across distant shots, and (c) fine-grained background details across intervening content.}
    \label{fig:consistency}
    \vspace{-10pt}
\end{figure}

To assess whether \name maintains the consistency of characters in long-range visual scenes when recurring subjects and details reappear after irrelevant video shots.  
Beyond aggregate inter-shot scores, we explicitly evaluate the character consistency across temporally distant shots. \cref{fig:consistency} illustrates that \name preserves recurring character identity and appearance across shot changes, non-human subject identity across distant shots, and fine-grained background details despite intervening content.

\section{Conclusion}\label{sec:conclusion}

We introduced \name, a multi-grid post-training framework that reformulates long-form multi-shot video generation as the joint generation of temporally ordered video chunks within a spatial grid. 
By distributing the full shot set across local temporal axes while jointly modeling all video chunks, \name reduces the number of shot transitions assigned to each temporal axis and overcomes the tendency of existing video generators to favor continuous motion over multi-shot realization. 
Built on MGLV and grid-aware post-training designs, \name achieves SoTA intra- and inter-shot consistency for subjects and backgrounds and directly generates \(1{,}616\)-frame multi-shot videos. 
Increasing the grid count from 16 to 64 extends a single generation to 6,464 frames under a fixed token budget. Building on this grid scaling, \name can further extend video length across successive generations by conditioning each new grid video on previous video chunks, showing that our method to be a scalable paradigm for long-form multi-shot video generation.

\section*{Acknowledgment}

This work is partially funded by an unrestricted gift from Google.
\clearpage
{\small\bibliographystyle{ieeenat_fullname}\bibliography{main}}
\clearpage
\appendix
\renewcommand\thesection{\Alph{section}}
\renewcommand\thefigure{S\arabic{figure}}
\renewcommand\thetable{S\arabic{table}}
\renewcommand\theequation{S\arabic{equation}}
\setcounter{figure}{0}
\setcounter{table}{0}
\setcounter{equation}{0}
\setcounter{page}{1}
\maketitlesupplementary

\section*{Appendix}

\section{Benchmark Detail}
\label{bmk}

We construct a story-level benchmark to evaluate long-form multi-shot video generation under the same narrative as \name. Unlike shot-oriented protocols that predefine individual shot prompts and shot boundaries, MGLV is constructed using detector-free fixed temporal intervals, where one chunk may contain one or multiple physical shots and the corresponding captions are aggregated into a character-aware Story Prompt. We therefore evaluate ordered event realization without assuming a one-to-one correspondence between textual descriptions, video chunks, and physical shots. Existing shot-oriented benchmarks such as ST-Bench~\cite{zhang2025storymem} evaluate a complementary setting and do not directly match this supervision granularity.

Our benchmark focuses on three key aspects:
\begin{itemize}
\item \textbf{Event coverage.} The generated video should visually realize all requested narrative events.
\item \textbf{Temporal progression.} The requested events should appear in the specified order and form a coherent story progression.
\item \textbf{Cross-event consistency.} Recurring characters, objects, and environments should remain visually consistent throughout the story.
\end{itemize}

In total, we curate 89 out-of-distribution stories with no narrative overlap with MGLV, spanning 5 visual domains: 3D CGI, anime, cinematic, realistic documentary, and stop motion. We further manually inspect all prompts to remove duplicate or highly similar stories and events.

\section{MGLV Details}

\begin{figure*}[t]
\centering
\begin{tcolorbox}[
colback=gray!4,
colframe=gray!35,
boxrule=0.5pt,
arc=2pt,
left=8pt,
right=8pt,
top=8pt,
bottom=8pt,
width=0.98\textwidth
]

\scriptsize

\textbf{Task.}
Split the input video into 10-second segments and return strict JSON only.

\textbf{Global entity catalog:}

Identify at most eight foreground, story-relevant main subjects across the
entire video and assign them stable identifiers
\texttt{C01}, \texttt{C02}, \ldots.

\begin{enumerate}[
leftmargin=1.8em,
itemsep=0.15em,
topsep=0.2em
]
\item \textbf{Subject relevance:}
An entity must be a foreground subject that a human annotator would likely
mention as the subject of a caption. Persistence alone is not sufficient.

\item \textbf{Entity grouping:}
Collapse visually similar people or animals appearing as a crowd, herd,
flock, pack, squad, audience, or group into one group entity. Do not assign
separate identifiers to interchangeable members unless the video consistently
distinguishes them as recurring individuals.

\item \textbf{Entity priority:}
Prioritize main humans, characters, and animals; then foreground vehicles,
creatures, and objects that move, are handled, or remain camera-centered.
Use a building, landmark, or prop only when the video is primarily about that
subject and no stronger foreground subject exists.

\item \textbf{Background exclusion:}
Ignore static scenery, generic environments, audience extras, and incidental
objects. Trees, grass, sky, clouds, roads, walls, floors, generic buildings,
room backgrounds, and generic furniture should not be treated as entities
unless they are the principal subject of the video.

\item \textbf{Entity appearance:}
Describe only static, identity-like visual attributes, such as species,
object type, color, clothing, hairstyle, accessories, and distinctive marks.
Do not include actions, states, or spatial relations in the appearance field.
\end{enumerate}

For each entity, provide:

\begin{itemize}[
leftmargin=1.8em,
itemsep=0.1em,
topsep=0.15em
]
\item \texttt{id}
\item \texttt{style}
\item \texttt{type}
\item \texttt{gender\_presentation}
\item \texttt{ethnicity\_or\_race}, only when the entity is human
\item \texttt{appearance}, containing concise static visual attributes
\end{itemize}

\textbf{Segment timeline:}

For every 10-second segment, provide its start time, end time, and the
identifiers of cataloged entities that are visibly present.

Use only visually observable information. Do not infer movie names,
source material, backstory, or internal states. Use \texttt{unclear} when
necessary. If no clear foreground subject exists, the entity catalog may be
empty. If no cataloged entity appears in a segment, return an empty
\texttt{present\_ids} list.

\textbf{Output format:}

\begin{quote}
\ttfamily
\{\\
\hspace*{1em}"entities": [\\
\hspace*{2em}\{\\
\hspace*{3em}"id": "C01",\\
\hspace*{3em}"style": "realistic/cartoon/anime/3d\_cgi/unclear",\\
\hspace*{3em}"type": "human/animal/object/creature/unclear",\\
\hspace*{3em}"gender\_presentation": "male/female/nonbinary/unclear",\\
\hspace*{3em}"ethnicity\_or\_race": "human only; otherwise unclear",\\
\hspace*{3em}"appearance": "concise static visual attributes"\\
\hspace*{2em}\}\\
\hspace*{1em}],\\
\hspace*{1em}"segments": [\\
\hspace*{2em}\{\\
\hspace*{3em}"start\_time": "HH:MM:SS",\\
\hspace*{3em}"end\_time": "HH:MM:SS",\\
\hspace*{3em}"present\_ids": ["C01", "C02"]\\
\hspace*{2em}\}\\
\hspace*{1em}]\\
\}
\end{quote}

Return only the JSON object.

\end{tcolorbox}

\caption{Qwen3-VL prompt template used to construct the global entity catalog
and coarse 10-second segment timeline for the MGLV dataset.}
\label{fig:mglv_global_prompt}
\end{figure*}

\begin{figure*}[t]
\centering
\begin{tcolorbox}[
colback=gray!4,
colframe=gray!35,
boxrule=0.5pt,
arc=2pt,
left=8pt,
right=8pt,
top=8pt,
bottom=8pt,
width=0.98\textwidth
]

\footnotesize

You are annotating a short video segment for dataset captions.

\textbf{Known entities:}

The following entity catalog is fixed. Do not repeat entity identifiers or
appearance descriptions in the scene or action fields.

\begin{quote}
\ttfamily
- C01: appearance=\{ENTITY\_01\_APPEARANCE\}\\
- C02: appearance=\{ENTITY\_02\_APPEARANCE\}\\
\hspace*{1em}\ldots
\end{quote}

\textbf{Requirements:}

\begin{enumerate}[
leftmargin=1.8em,
itemsep=0.15em,
topsep=0.2em
]
\item \textbf{Scene:}
Describe only the visible location, setting, and relevant objects. Use a noun
phrase of at most 12 words. Do not include actions or entity identifiers.

\item \textbf{Action:}
Describe only the visible action using a verb phrase of at most 10 words.
Do not include a grammatical subject or entity identifiers.

\item \textbf{Visual style:}
Select exactly one label from
\texttt{realistic}, \texttt{cinematic}, \texttt{comic},
\texttt{cartoon}, \texttt{anime}, \texttt{3d\_cgi},
\texttt{stop\_motion}, \texttt{pixel\_art},
\texttt{game\_render}, or \texttt{unclear}.

\item \textbf{Video style:}
Specify the shot type, camera motion, lighting condition, and overall tone
using only the allowed labels in the output schema.
\end{enumerate}

Be concise and avoid repetition. Describe only visually observable facts.
Do not infer backstory or unobservable information. Use \texttt{unclear}
whenever the visual evidence is insufficient.

\textbf{Output format:}

\begin{quote}
\ttfamily
\{\\
\hspace*{1em}"scene": "noun phrase, at most 12 words",\\
\hspace*{1em}"action": "verb phrase, at most 10 words",\\
\hspace*{1em}"visual\_style": "one allowed visual-style label",\\
\hspace*{1em}"video\_style": \{\\
\hspace*{2em}"shot\_type":
"close-up/medium/long/over-the-shoulder/unclear",\\
\hspace*{2em}"camera\_motion":
"static/pan/tilt/handheld/zoom/dolly/unclear",\\
\hspace*{2em}"lighting":
"daylight/indoor\_warm/night/low\_key/backlit/unclear",\\
\hspace*{2em}"tone":
"cinematic/home-video/dramatic/comedic/neutral/unclear"\\
\hspace*{1em}\}\\
\}
\end{quote}

Return only the JSON object.

\end{tcolorbox}

\caption{Qwen3-VL prompt template used to generate structured captions for
individual MGLV subvideo segments.}
\label{fig:mglv_segment_prompt}
\end{figure*}

We employ Qwen3VL 8B~\cite{Qwen3-VL} in a two-stage \textit{Character-Aware Story Annotation} pipeline. Given a temporally concatenated subvideo segment, the first stage identifies up to eight recurring foreground entities, assigns persistent identifiers, and records their visibility over consecutive 10-second intervals. Conditioned on this global entity catalog, the second stage annotates the scene, action, visual style, and cinematographic attributes of each interval (see \cref{fig:mglv_global_prompt}). 

As shown in \cref{fig:mglv_segment_prompt}, in the second stage, we convert the structured outputs into a \textit{Story Prompt} by associating recurring entities with persistent special tokens, such as \texttt{<C01>}, composing the interval-level descriptions, and concatenating them in temporal order. The special prefix \texttt{<grid N>} declares the target grid layout, where \(N\) denotes the number of video chunks. The annotation intervals provide ordered narrative supervision but are not required to align one-to-one with grids or physical shots. We sample videos at 0.5 FPS, use deterministic decoding, and automatically validate and retry malformed JSON outputs.

% \section{Human Study}

\section{More Results}

\begin{figure*}[t] 
    \centering
    \includegraphics[width=\textwidth]{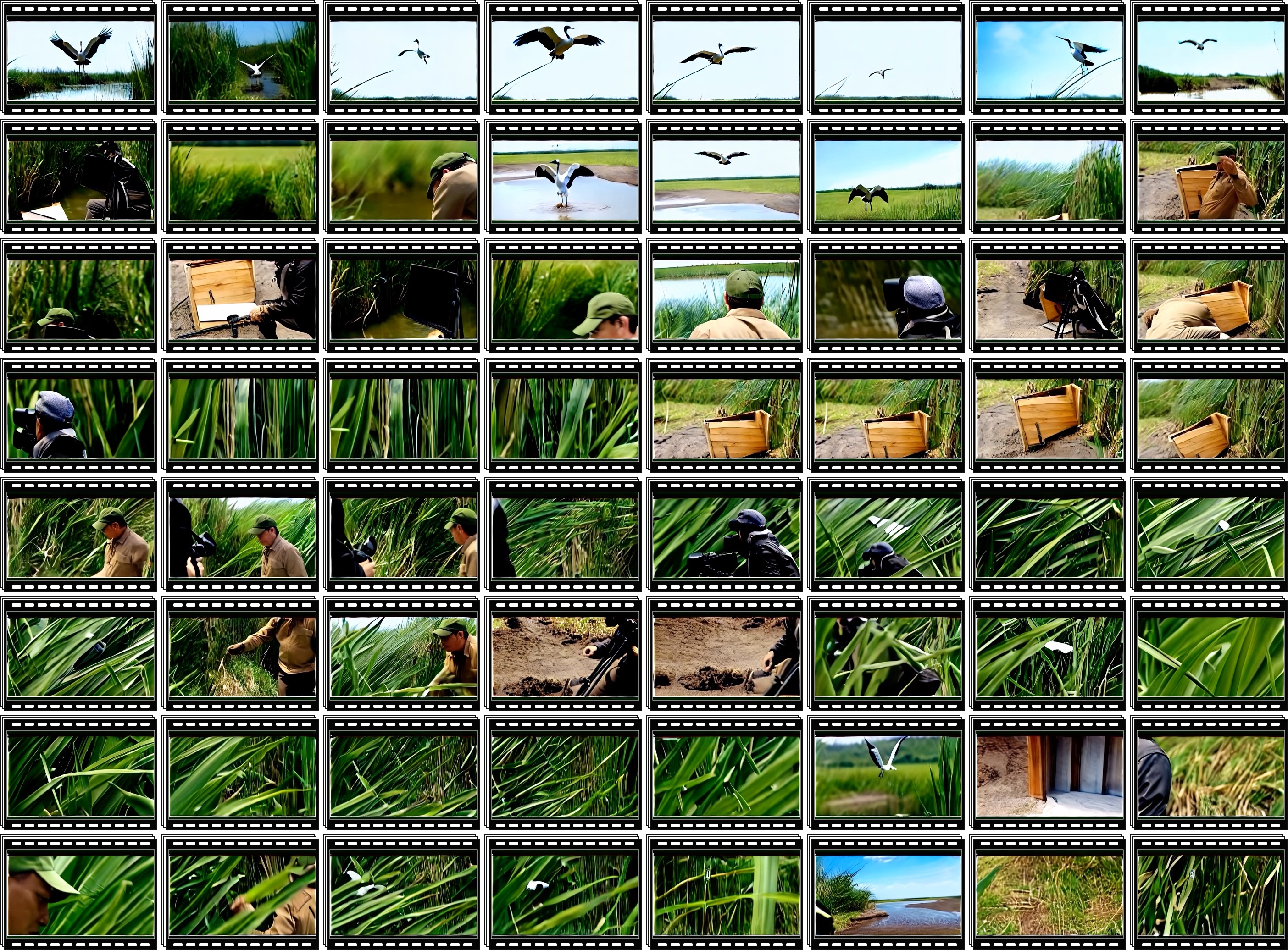} 
    \caption{Qualitative results for a 64-grid.} 
    \vspace{-2ex} 
    \label{fig:commercial} 
\end{figure*}

Figure~\ref{fig:commercial} presents an additional qualitative result from \name-64, which jointly generates an \(8\times8\) grid of temporally ordered video chunks and produces a \(6{,}464\)-frame multi-shot video after unpacking. The extended sequence covers diverse events and viewpoints while preserving recurring subjects and coherent visual context, further demonstrating the scalability of the proposed grid representation.

% \section{Limitation}

% \section{Future Work}

\end{document}